\documentclass[fleqn,10pt]{wlscirep}
\usepackage[utf8]{inputenc}
\usepackage{psfrag}
\usepackage{subfig}
\title{Automatic Classification of Cancerous Tissue in Laserendomicroscopy Images of the Oral Cavity using Deep Learning}

\author[1,*]{Marc Aubreville}
\author[2,3]{Christian Knipfer}
\author[3,4]{Nicolai Oetter}
\author[1]{Christian Jaremenko}
\author[5]{Erik Rodner}
\author[5]{Joachim Denzler}
\author[6]{Christopher Bohr}
\author[7,3]{Helmut Neumann}
\author[4,3]{Florian Stelzle}
\author[1,3]{Andreas Maier}
\affil[1]{Pattern Recognition Lab, Computer Science, Friedrich-Alexander-Universit{\"a}t Erlangen-N{\"u}rnberg}
\affil[2]{Department of Oral and Maxillofacial Surgery, University Medical Center Hamburg-Eppendorf}
\affil[3]{Erlangen Graduate School in Advanced Optical Technologies (SAOT), Friedrich-Alexander-Universit{\"a}t Erlangen-N{\"u}rnberg}
\affil[4]{Department of Oral and
  Maxillofacial Surgery, University Medical Center Erlangen, Friedrich-Alexander- Universit{\"a}t
  Erlangen-N{\"u}rnberg}
\affil[5]{Computer Vision Group, Friedrich-Schiller-Universit{\"a}t Jena, Germany}
\affil[6]{Department of Otorhinolaryngology, Head and Neck Surgery, University Medical Center Erlangen, Friedrich-Alexander-Universit{\"a}t Erlangen-N{\"u}rnberg}
\affil[7]{First Department of Internal Medicine, University Medical Center Mainz, Johannes Gutenberg-Universit{\"a}t Mainz}

\affil[*]{marc.aubreville@fau.de}

\begin{abstract}
Oral Squamous Cell Carcinoma (OSCC) is a common type of cancer of the oral epithelium. Despite their high impact on mortality, sufficient screening methods for early diagnosis of OSCC often lack accuracy and thus OSCCs are mostly diagnosed at a late stage. Early detection and accurate outline estimation of OSCCs would lead to a better curative outcome and an reduction in recurrence rates after surgical treatment.

Confocal Laser Endomicroscopy (CLE) records sub-surface micro-anatomical images for in vivo cell structure analysis. Recent CLE studies showed great prospects for a reliable, real-time ultrastructural imaging of OSCC in situ.

We present and evaluate a novel automatic approach for a highly accurate OSCC diagnosis using deep learning technologies on CLE images. The method is compared against textural feature-based machine learning approaches that represent the current state of the art.

For this work, CLE image sequences (7894 images) from patients diagnosed with OSCC were obtained from 4 specific locations in the oral cavity, including the OSCC lesion. The present approach is found to outperform the state of the art in CLE image recognition with an area under the curve (AUC) of 0.96 and a mean accuracy of 88.3\,\% (sensitivity 86.6\,\%, specificity 90\,\%).

\end{abstract}
\begin{document}

\flushbottom
\maketitle
%
%
\thispagestyle{empty}

\section{Introduction}
%
%
%
%

Squamous Cell Carcinoma is a form of cancer that originates from
squamous cells of the skin or mucous membranes. In the area of the
head and neck, the malignant transformation of these cells leads to a
worldwide incidence of 1.3 million new cancer cases per year \cite{Forastiere:2009bw,Ferlay:2014ht}. Most cases of Head and Neck Squamous Cell
Carcinoma (HNSCC) are already at
an advanced stage when diagnosed which significantly reduces the survival rate after curative treatment\cite{Muto:2004hy}.
The gold standard for diagnosis of HNSCC is
an invasive biopsy of the lesion, followed by a histopathological assessment\cite{Swinson:2006bj}. In addition imaging methods such
as narrow band imaging\cite{Muto:2004hy} and Raman spectroscopy
\cite{Knipfer:2014jm,Swinson:2006bj} are considered as emerging tools used for non-invasive detection of malign neoplasms of the head and neck. Recently, Confocal Laser Endomicroscopy (CLE)
\cite{Laemmel:2004fd}, an imaging technique that has been widely used and validated in pathological tissue diagnosis of the gastrointestinal tract
\cite{Neumann:2010hb,Hoffman:2006ei}, has also been studied for its potential of reliably diagnosing HNSCC in
situ\cite{Oetter:2016cp,Nathan:2014ky}. 

Compared to bright light endomicroscopy, CLE does not only have the
advantage of an exceptionally high magnification of up to
1000x\cite{Oetter:2016cp}, but also provides a better depth
penetration \cite{Helmchen:2002vt}, allowing for diagnosis of
  malignancies approximately 100 microns below the surface.  For this imaging technology, a 
fiber bundle that is connected to a laser source in the cyan spectrum ($488\,$nm) is
applied on biological tissue in cavities of the human body. A contrast agent (fluorescein) is administered
to the patient by i.v. injection prior to the examination. This agent accumulates in
the intercellular gaps and emits light (at $520\,$nm) upon excitation by the laser light, thus enabling 
imaging of cell outlines. The beam path, including laser source and a pinhole are
constructed in such a way that light reflected from outside the focal
plane is geometrically eliminated \cite{SCA:SCA4950100403,Abbaci:2014wn}. As both, the detector and the laser source,
are in the same focal plane, the system is called `confocal' \cite{Hoffman:2006ei}.

Since the grayscale images of biological cells in its compound as acquired by CLE are unlike other imaging
techniques, special training for the pathologist or surgeon
interpreting the images is of great importance \cite{Abbaci:2014wn},
and examiner's experience has a distinct influence on the over-all CLE
performance \cite{Neumann:2011tc}. Thus, an automatic detection of
HNSCC in micro-anatomical imaging techniques (such as CLE) could be a valuable tool for screening without
the need of subjective image interpretation and extensive training in image interpretation. With an
increased availability of CLE imaging devices, an examiner-independent screening  of
suspicious tissue could help diagnosing HNSCC at earlier stages and
thus improve treatment outcomes.

Besides the diagnosis of cancer, another important field of application of an automatic classification in CLE imaging is the surgical therapy of the malignancy in terms of computer-aided surgery. Finding an adequate resection margin of a
tumor is crucial for the overall success of the curative
therapy. Failure to find this margin with a subsequent recurrence of the cancer is the most common cause of
death for patients with HNSCC \cite{Nathan:2014ky}. Since CLE images
are taken at a greater depth than common endoscopic images, a larger
field involvement beneath the surface can be detected, which has the
potential to reduce morbidity and mortality after surgery \cite{Nathan:2014ky}.

\subsection{Automatic Classification of CLE images}

The image sequences acquired using CLE imaging are very different to other
forms of medical images, as they display a small horizontal layer,
up to $100\,\mu$m beneath the surface of the probe \cite{Mennone:2017jj}. For the
present study, images were acquired using a standalone probe-based CLE
system (Cellvizio, Mauna Kea Technologies, Paris, France). The probes used for imaging were CystoFlex UHD R and ColoFlex UHD, both having a similar field of view and penetration depth. The
inspected area is approximately $250\,\mu $m in width and height, with
a total number of 576 px. Automatic classification
using CLE imaging has already been proven to show valid results in
clinical studies \cite{Andre:2012kw,Kamen:2016jw,Veronese:2013hb}. However, for each anatomical
location, CLE images differ as the tissue under observation also differs.  Andr\'{e} \textit{et al.} have shown that automatic image recognition
using probe-based CLE can be used successfully
for detection of neoplastic tissue in the colon region \cite{Andre:2012kw}. Their results
indicate that automatic recognition can yield similar results to the
diagnosis by endoscopy experts. Kamen \textit{et al.} have successfully applied machine learning techniques on CLE images of brain tumors. \cite{Kamen:2016jw}

Jaremenko \textit{et al.} have first employed automatic image recognition on CLE images
of the oral cavity, using the classical pattern recognition
workflow with a number of textural features (local binary pattern
(LBP), gray-level coocurrence matrix (GLCM) or local histogram
statistics) and subsequent machine
learning techniques (random forest (RF) and support vector machine)
for the classification\cite{Jaremenko:2015kh}. 
Dittberner \cite{Dittberner:2016jv} and Rodner \cite{Rodner:2015ts}
have shown that also segmentation-based methods have the
potential to be applied to cancer recognition in CLE images of the
head and neck region.  They extracted the cell borders from the image
and used a distance transform with successive histogram calculation in
order to use the cell size as a feature for classification. 
On the respective data set, they reached a mean cross-validation accuracy of
74\,\%\cite{Dittberner:2016jv}. 
In both cases, the number of images used in the
recognition task was rather limited, calling for validation of the
techniques with a substantial increase in image material. 

The present study features a rather large amount of data, enabling
a different class of machine learning techniques: deep
artificial neural networks (DNN).
While the classical, feature-based
approach incorporates prior knowledge about the classification task
(by proper selection of features), deep learning techniques commonly
are solely calculated on the raw input data. The number of unknown
parameters is much higher for a DNN approach, compared to the
classical workflow, which also calls for a much higher amount of
training data.


\subsection{Convolutional Neural Nets}
For image recognition, one
particularly successfully used method in recent years is the
application of convolutional layers in these deep networks. Convolutional
filters are inspired by the pattern recognition of the
visual cortex, where so-called receptive fields show activations based
on distinct spatial patterns of the visual scene \cite{Hubel:1968ep}. 


Methods employing Convolutional Neural Networks (CNNs) have won all
major image recognition challenges (like the ILSVRC challenge
\cite{Russakovsky:2015bu}) and have recently also been successfully
applied in the field of medical image analysis
\cite{Shin:2016cx,Roth:2016eg,Esteva:2017ct,Litjens:2016fr} and even reconstruction \cite{Wurfl:2016jq}. 


\section{Material}
For the present study, $N=116$ video sequences from $12$ patients with cancer in
the oral cavity were acquired at the Department of Oral and Maxillofacial Surgery (University Hospital Erlangen). Written consent was obtained from all patients and institutional review board approval
was provided prior to the study (ethics committee of the University of Erlangen-N{\"u}rnberg; reference number: 243\_12 B).
 
\begin{figure}[!t]
\centering
\subfloat[ ]{
\includegraphics[height=2.0in]{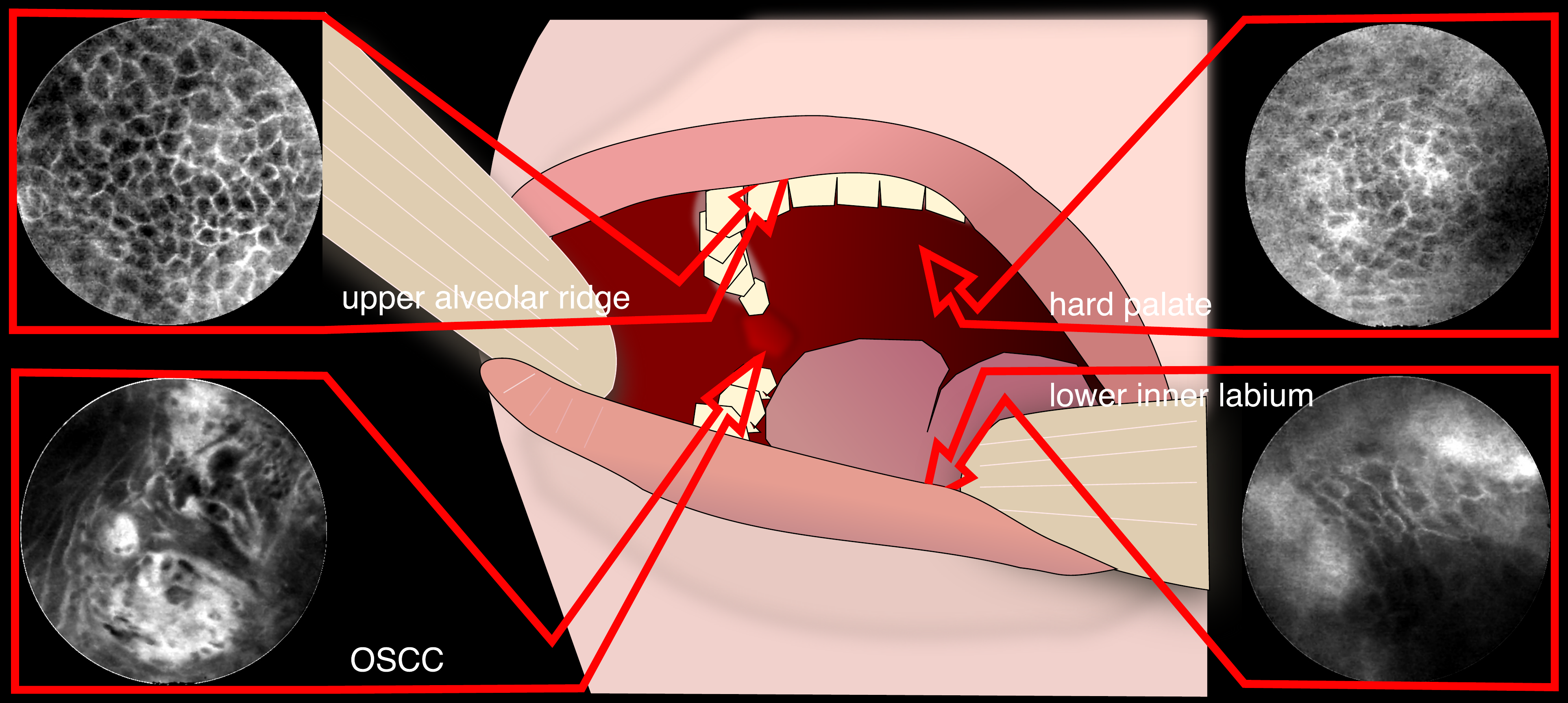}
\label{fig:Regions}
}
\subfloat[ ]{
\includegraphics[height=2.0in]{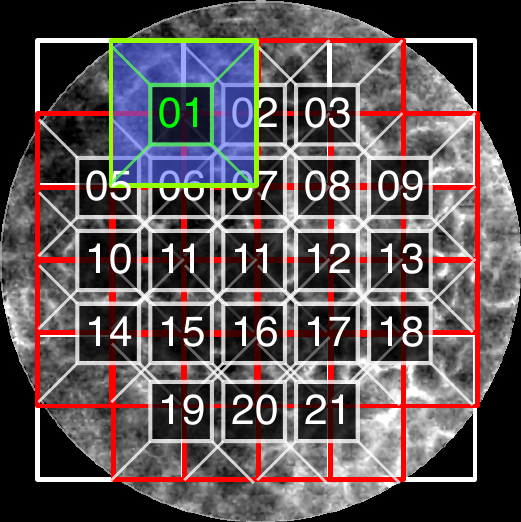}
\label{fig:imagePatches}
}
\caption{Left: CLE recording locations. Additionally, the region of the
  suspected HNSCC was recorded. Right: Division of (resized) image into patches of size 80x80 px. Only
  patches that were inside the image mask and had no artifact labels
  within them were considered for classification.}
\end{figure} 

From each patient, image sequences from the suspected carcinogenic region
were recorded. Additionally, images from three other (physiological) regions were made: From the inner lower
labium, the upper alveolar ridge and the region of the hard palate (see
Fig. \ref{fig:Regions} and table \ref{tab:locations}). Specimen from all tumorous regions were resected after image acquisition and
histologically verified by a trained pathologist. The video
sequences acquired before surgery, were hand-cut
by a clinician expert in order to remove parts where the instrument was not
properly placed or did not show the tissue to be investigated. This resulted in
approximately 11.000 images, having different image qualities and some
impaired by heavy artifacts.

The most common artifact (2659 images) in the data set was noise,
ranging from slight added noise to images containing only noise.
This may be related to an illumination problem, where no contrast
agent is located under the probe, or the probe is not properly placed
on the mucosa. 
Another common (1455 images) artifact is motion artifacts, originating from movement of the probe during image acquisition, resulting in shearing, compression or elongation of the image or parts of the image. This effect severely deteriorates the image, which is why affected images were also excluded. 
After also excluding images with optical artifacts (such as mucus or blood drops on the probe) and images of otherwise bad quality, 7894 
 images of good quality remain for the purpose of image recognition. This results in a mean image count of 658 images per patient ($\sigma=399$). All images were either assigned to the class ``clinically normal'' or the class ``carcinogenic'', with an almost even distribution of both classes (see table \ref{tab:locations}).

\begin{table*}\centering\begin{tabular}{l|l|c|c|c}
  Class & location & No. total & No. good images & Percentage in final data set  \\
  \hline
  normal & alveolar ridge & 2133 & 1951 & 24.71\,\%\\
  normal & inner labium & 1327 & 1317 & 16.68\,\% \\
  normal & hard palate & 955 & 811 & 10.27\,\% \\
  carcinogenic & various & 6530 & 3815 & 48.33\,\% \\
\end{tabular}
\caption{Number of images of different regions}
\label{tab:locations}
\end{table*}

\section{Methods}

\subsection{Patch-extraction of images}

In principal, we follow the workflow of Jaremenko \textit{et al.} \cite{Jaremenko:2015kh} in that images are divided into patches, where information is extracted and dimensionality reduced, and subsequent fusion of the information to achieve classification per image.
 
The images were extracted from the raw data container the CLE imaging
system produces in a 16 bit grayscale format \cite{Aubreville:2016}.
Since CLE images are often noisy, we propose to reduce processing complexity and noise in the image by scaling the image down to half the size. This way also more relevant structures are captured in a single patch.  

Each CLE image has a size of 576x576 px.\footnote{Due to
  manufacturing variances and subsequent calibration, the images
  actually have a width and/or height of 576px, 578px or 580px.}
The images have a circular shape which makes processing the whole
image at a time difficult. Because of this, we're dividing the resized
image (denoted $\textbf{I}$)  into
patches (denoted $\textbf{P}$) of size 80x80 px with an 50\,\% overlap, centered
around the middle of the image, resulting in 21
patches out of 1 image (see Fig. \ref{fig:imagePatches}):
\begin{equation}
 \textbf{I} \xrightarrow{\textrm{patch extraction}} \left[{
    \textbf{P}_0, \textbf{P}_1, \dots \textbf{P}_{N_\textbf{I}}}
    \right] \hspace{0.3in} N_\textbf{I} \leq 21 
\end{equation}

Each resulting patch $P_n$ is assigned a coordinate quadruple
that delimits the corners of the patch:
\begin{equation}
\vec{c}(\textbf{P}_i)  = \left[ c_{1,i}, c_{2,i}, c_{3,i}, c_{4,i}\right]
\end{equation}
where $(c_1,c_3)$ is the left top corner and $(c_2,c_4)$ the bottom right corner.

Within the images with overall good image quality, a number of images have minor known, annotated
artifacts (annotated as rectangles within the image), only affecting the image slightly. Patches with artifacts are removed from the image
recognition task, while the rest of said images is included. This means, however, that the number of patches per
image is not constant, and thus restricts the possibilities for
whole-image classification.

To assume no prior knowledge about illumination of the image, all
patches were whitened using a standard scaling to achieve zero mean and
unity standard deviation. 

\subsection{Data augmentation for training}
Since CLE images have no natural orientation, a rotated CLE image is
still a valid image. Because of this, we enrich the data provided to
the classifier by arbitrarily, randomly rotated copies of known images. These
augmented images however may not be used for testing the algorithm,
since we can't completely eliminate the possibility that the images
have some inherent properties that are indeed rotation-variant, for
example originating from a common hand position of the physician.  We
used a 2-fold augmentation, meaning that out of each original image, two randomly rotated
 copies were created and fed to patch extraction. 

In order to avoid introducing bias for one of the classes, each classifier receives an equal
distribution of both classes for training by removing augmented images of the majority class. 

\subsection{Classification approaches}
\vspace{0.5em}
\subsubsection{Textural feature-based classification}
The approach described by Jaremenko \textit{et al.}\cite{Jaremenko:2015kh} uses
different textural features. Amongst the best scoring were features
based on local binary
patterns (LBP) and gray-level co-occurance matrices (GLCM), which is why these were
included for comparison. 

LBPs describe a pixel gray value in
relationship to its neighboring pixels and were successfully used for
image recognition tasks such as face recognition \cite{Ahonen:2006gr}
 or cell phenotype classification \cite{Nanni:2010eb}. Jaremenko et
 al. use rotation invariant uniform LBPs and calculate a histogram for
 each patch of these. Instead of using the histograms themselves as
 features for the classifier, the mean and standard deviation of the
 features over all patches of an image are used.

GLCMs, on the other side, describe the statistical occurrence of
certain gray values in neighboring pixels within an image
patch. From these matrices, certain features that characterize properties of
the image can be calculated. Jaremenko \textit{et al.} use the GLCM-based
features described by Haralick
 \cite{Haralick:1973bs}, as
well as the extended features described by Baraldi \textit{et al.} 
\cite{Baraldi:1995js}.

As classification approach, support vector machine (SVM) and random
forest (RF) were used.

\begin{figure*}[!t]
\centering
\includegraphics[width=\textwidth]{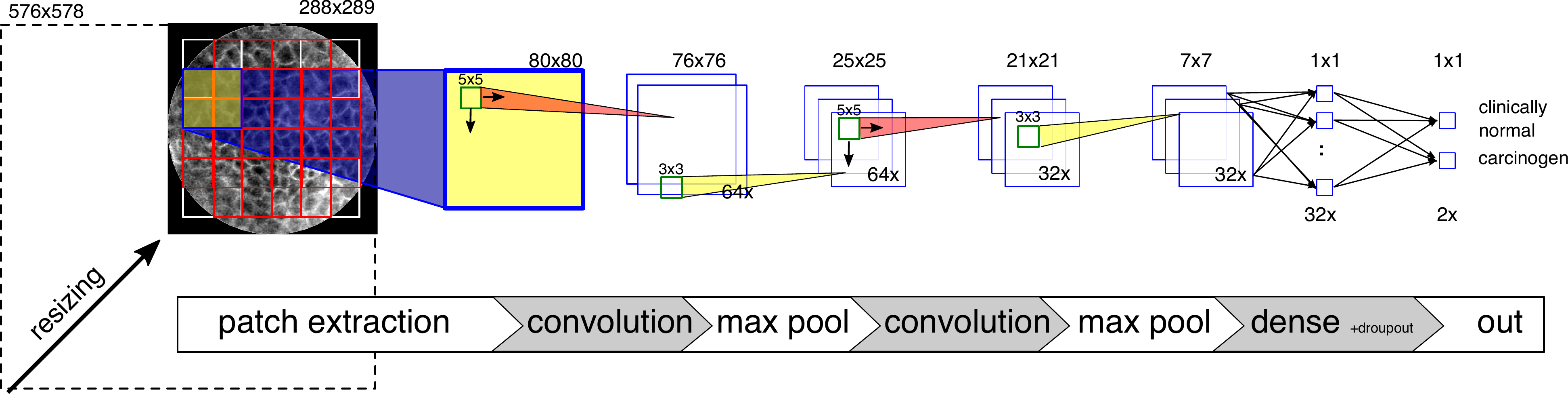}
\hfil
\caption{Overview of the CNN-based patch extraction and classification.}
\label{CNNtoolchain}
\end{figure*}

Jaremenko \textit{et al.} reported very convincing results, especially for
GLCMs with SVM classifier 
(accuracy$\,=\,99.2$\,\%) and also good values for
LBPs (accuracy$\,=\,91.2$\,\%), on a small database of only 251 images
\cite{Jaremenko:2015kh}, however. GLCM-based features were calculated with
different image-level configurations (8,\,16 and 32 levels), and showed
similar results. Since generalization of these results can't
be assumed, we included both the GLCM-features and the LBP-features in
our evaluation. The approaches were evaluated on patch sizes of 80x80 px and 105x105 px,
with comparable results. Vo \textit{et al.} re-evaluated both feature sets on a
much larger database of vocal cords CLE images \cite{Vo:2016} and
found comparable results for GLCM-features and LBP-features, with LBPs
performing slightly better and little difference amongst the different
configurations of features (i.e. image levels for GLCMs). 


We included the following configurations for comparison:
\begin{enumerate}
\item \textbf{RF-LBP@1.0x} Random Forest-classified result using the LBP
  feature set (radii = [1,3,5], number of neighbors = [8,16,24], rotation invariant uniform LBPs, mean and std over all patches, number of trees=500)
\item \textbf{RF-LBP@0.5x} equal to RF-GLCM@1.0x, but calculated on a
  resized (factor 0.5) image
\item \textbf{RF-GLCM@1.0x} Random Forest-classified result using the GLCM-based
  feature set with 16 image levels (mean and std over all patches, number of trees=500)
\item \textbf{RF-GLCM@0.5x} equal to RF-GLCM@1.0x, but calculated on a
  resized (factor 0.5) image
\end{enumerate}

\vspace{0.5em}
\subsubsection{Patch-based Convolutional Net Processing}

Convolutional Neural Networks (CNNs) do not rely on feature extraction as
a first step, but take an image as input and have feature extraction
inherently within the network. 

In principal, CNN machine learning can be run on the whole image as well as
on patches. If patches can be considered a representative sample of
the whole image, patch extraction is a beneficial approach because of the following reasons:
\begin{itemize}
\item Classification of patches reduces the order of the pattern
  recognition problem. As the number of parameters to be learned for the
  pattern recognition algorithm, in our case the neural network, goes
  quadratically with the image length and width, it is
  dramatically reduced. Since CNN approaches in general require a
  large amount of data in comparison with feature-based machine learning
  approaches, this is an important factor.
\item Reduction of classification error. If an independent error is
  assumed on the result of the classification, fusion of the single
  patch classification results will reduce the overall error of the
  image classification.
\end{itemize}
In our case, we consider the whole image \textit{cancerous} or
\textit{clinically normal}, since no sub-image labelling was performed and it was observed that the vast majority of patches show the same characteristics as the image classification.

Our convolutional network is based on the LeNet-5 network proposed by Lecun
\cite{LeCun:1998fv}: A convolutional layer with 64 filters of size 5x5 px is followed
by a max-pooling layer (3x3 px), another convolutional layer with 32
filters of size 5x5 px, another max-pooling layer (3x3 px), one fully connected layer with drop-out and an output
layer (see Fig. \ref{CNNtoolchain}). The network was trained with
the TensorFlow framework \cite{Abadi:2016vn} at a
learning rate of 0.001 using the Adam optimizer \cite{Kingma:2014us} for cross-entropy minimization. The network has 103170 learnable parameters and was trained completely from scratch.

The convolutional net $n$ assigns to each patch $\textbf{P}$ an a-posteriori probability $\vec{p}$ for a given class $c$ ($c=1$ is cancerous, $c=0$ is normal):
\begin{equation}
n: \mathbb{R}^2 \rightarrow \mathbb{R}, n(\textbf{P}) \rightarrow
\left[ \begin{matrix}p({c=0}) \\ p({c=1}) \end{matrix} \right]
\end{equation}

This probability can be mapped on the image again. For this, we define
first of all the area function of a patch $\textbf{P}_i$:
\begin{equation}
A_{x, y}(\mathbf{P_i}) = \begin{cases}1 & \text{if } (x, y) \in [c_{1}, c_{2}] \times [c_{3}, c_{4}]  \\
0 & \text{else}
\end{cases}
\end{equation}

From this we derive the patch activity map:

\begin{equation}
\mathrm{PA}_{x,y} = \left( \sum_i A _{x,y} (\textbf{P}_i) \right) \geq 1
\end{equation}

and the patch count map:
\begin{equation}
\mathrm{PC}_{x,y} = \max \left( 1, \sum_i A _{x,y} ( \textbf{P}_i ) \right)
\end{equation}

These maps are constant and geometrically point-symmetric to the
center in case no artifact is available in the image. Finally, the probability map of the image is:
\begin{equation}
\mathrm{PM}_{x,y} = \mathrm{PA}_{x,y} \cdot \mathrm{PC}_{x,y}^{-1} \cdot\sum{ A_{x,y}(\textbf{P}_i) \cdot p_i(c=1)} 
\end{equation}

Since the classification approaches are balanced, we neglected the
probability of the clinically normal class ($c=0$) for this classification problem because the information
is contained in the probability of the cancerous class ($c=1$).

From this we derive a scalar probability number for the image \textbf{I}:
\begin{equation}
     p(\mathbf{I}) = \left( \sum_{x,y} \mathrm{PA}_{x,y} \right)^{-1}
     \sum_{x,y} \mathrm{PM}_{x,y}
    \label{eqn:ppf}
\end{equation}




The a posteriori probabilities of the image are thus fused into a
single probability number, thus we denote the approach \textbf{patch probability
  fusion (ppf)} method. 


\vspace{0.5em}
\subsubsection{Whole image classification using Transfer Learning with CNNs}
\label{CNNimageDet}

Besides the patch-based detection of images, it is also possible to feed the complete image to the classification method. While the number of images fed to the classification training decreases, the network complexity and thus the number of free parameters increases dramatically for this approach, fueling the need for a regularization in order to prevent the network from overfitting.

\begin{figure*}[!t]
\centering
\includegraphics[width=\textwidth]{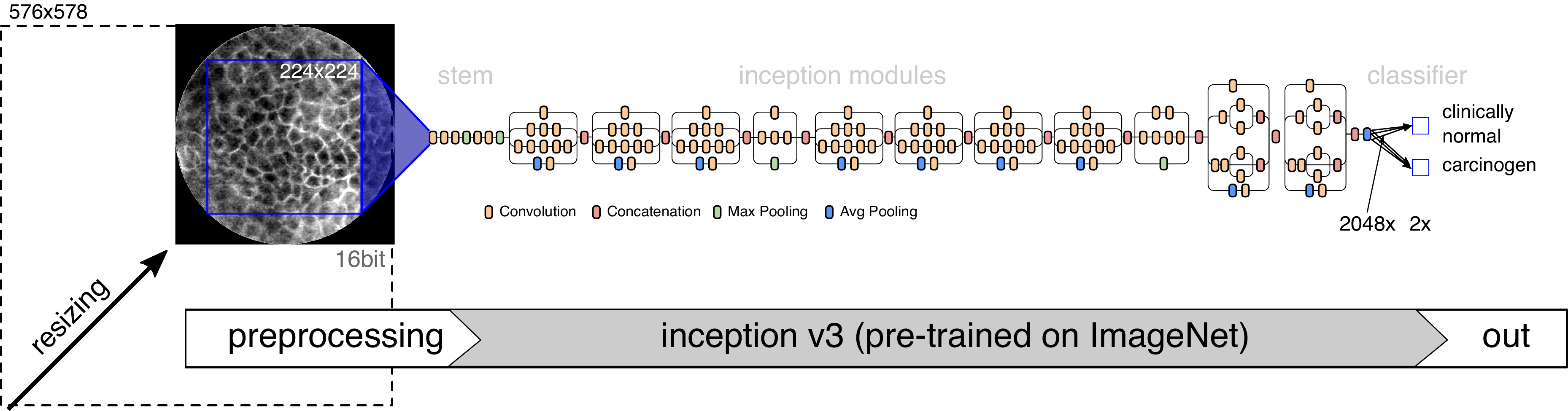}
\hfil
\caption{Overview of the transfer learning approach, based on Szegedy's Inception v3 \cite{Szegedy:2016cv}, pre-trained on the ImageNet database \cite{imagenet_cvpr09} }
\label{CNNTFtoolchain}
\end{figure*}

Commonly, this problem is solved using network architectures, that were pre-trained on images of a different domain (e.g. real-world photography images or other medical images) and are then fine-tuned on a new image data set (transfer learning) \cite{Shin:cx,Esteva:2017ct}. We use the Inception v3 network from Szegedy et al.\cite{Szegedy:2016cv}, pre-trained using ImageNet \cite{imagenet_cvpr09}, and replace the final dense layer and softmax layer with a new two node dense layer and subsequent softmax layer (see figure \ref{CNNTFtoolchain}).  

Since CLE image data is 16 bit at a single wave length and the Inception v3 takes 8 bit RGB images, some pre-processing needs to be applied.
In order to reduce 16 bit depth into 8 bit, we apply a dynamic compression: The image is scaled according to the following percentile scaling rule:

$$ \mathbf{I}_\textrm{8\,bit} = \frac{255}{\mathit{P}_{99.5\%} - \mathit{P}_{0.5\%}} \cdot (I-P_{0.5\%})$$

with $\mathit{P}_y$ being the $y$th percentile of pixel intensity values within the circular view area.

The resulting 8\,bit image is then mapped to a greyscale RGB image, from which the maximum square area is extracted. It is defined as a square with dimensions $w=h=\frac{2}{\sqrt{2}} \cdot r_\mathrm{CLE}$ around the
center of the image, where
$r_\mathrm{CLE}$ is the radius of the circular CLE view area in pixels.

In order to fit the target input dimension of 224x224 pixels of the Inception v3 network, a final pre-scaling of approximately 0.55x is applied. For this task, too, data augmentation was applied during training. In this case, a random rotation was applied to the image, before cropping the maximum square image around the center. Other augmentation methods like arbitrary scaling have not been applied, because of absolute dimensions of the medical images. 

Each network in the cross-validation was trained for 3000 epochs of 100 steps, using the Adam optimizer with a step size of 0.01 for the new layers and no adaptation for the layers taken from the Inception v3 network.

\subsubsection*{Evaluated approaches and configurations}

In total, we evaluated three CNN-based approaches
\begin{enumerate}
\item \textbf{CNN/ppf@0.5x} CNN-based detection using patch
  probability fusion, patch size 80x80 px, resized image (scaling factor 0.5) (see figure
  \ref{CNNtoolchain})
\item \textbf{CNN/ppf@1.0x} CNN-based detection using patch
  probability fusion, patch size 80x80 px, original
  (unscaled) image
\item \textbf{CNN/TF@0.55x} Transfer learning approach using pre-trained CNNs, maximum square image, scaled to 224x224
\end{enumerate}

The code for all approaches is publicly available at the following URL: \textbf{filled in after paper acceptance}.

\section{Results}

\subsection{Cross-Validation}

We evaluated both the feature- and the CNN-based methods using a leave-one-patient-out
cross-validation, i.e. one patient always represented the
test data and all others the training data. This way, inherent
correlation within the image sequences (as these were recorded as
videos) did not play a role in the evaluation. All classifications
results from this validation are subsequently concatenated to a final
result vector to ease comparison of the evaluated methods. 

\subsection{Textural feature-based Classification}

The method by Jaremenko \textit{et al.}\cite{Jaremenko:2015kh}, using textural
features on image patches,
yielded cross-validation accuracy ratings of  77.9\,\% (sensitivity: 80.2\,\% , specificity: 72.2\,\%) on the LBP-based feature vector and 
70.6\,\% (sensitivity: 75.5\,\%, specificity: 63.9\,\%) on the GLCM-based feature
vector, both using random forest classifiers. This performance is
significantly different to the original publication. The reason for
this is that the image quality in the present data set is much more
mixed as more patients and more anatomical locations were considered.

\begin{figure}
\centering
\includegraphics[width=3.0in]{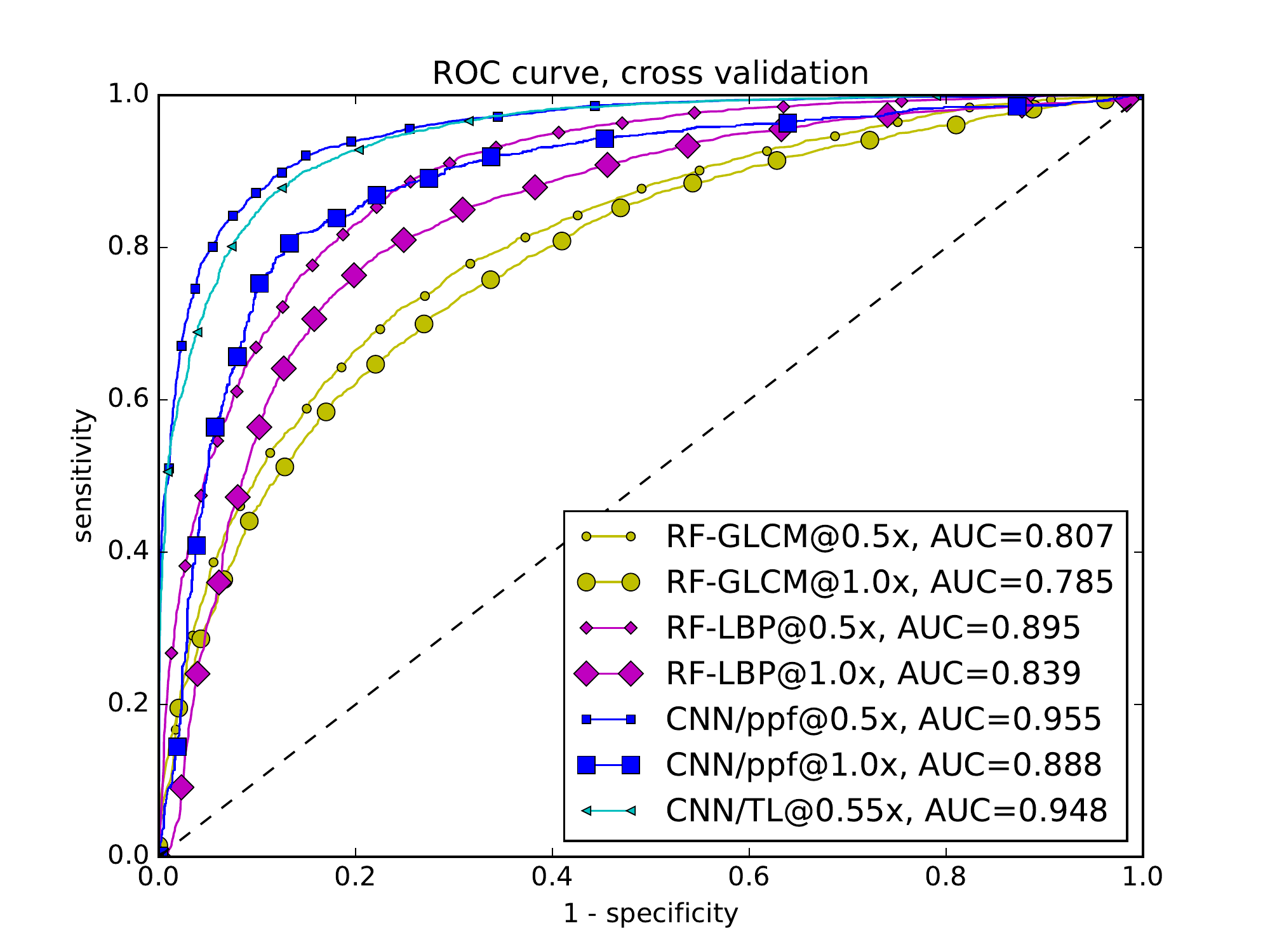}
\caption{ROC curve of cross-validation. All results of the single
cross-validation steps were combined into one result vector. }
\label{fig:roc}
\end{figure}

For both approaches,
resizing the image to half the original dimensions (factor 0.5)
before patch extraction yielded significant advances in detection
performance: The LBP-based classifier improved to an accuracy of
81.4\,\% (sensitivity: 84.7\,\%, specificity: 78.2\,\%) and the
GLCM-based classifier to an accuracy of 73.1\,\% (sensitivity: 77.5\,\%,
specificity: 69.5\,\%). The receiver operating characteristic (ROC) curve evaluation in figure
\ref{fig:roc} shows the sensitivity and specificity for different discrimination thresholds. In result, the Area Under Curve (AUC) improved for the scaled
patches and LBP features from 0.84 to 0.90, and for GLCM-features from
0.78 to 0.81.

\subsection{CNN-based approaches}

\subsubsection*{Patch-probability fusion}

The CNN training was performed on 40 batches, each containing around 12.000
patches. Convergence was reached usually after around 200 epochs, yielding
patch-training accuracies of between 83\,\% and 88\,\%. The cross-validation accuracy as evaluated on patches was 78.47\,\%
(sensitivity=76.31\,\%, specificity=80.42\,\%).

Fusion of patch a posteriori 
probabilities to image probabilities as defined by the patch probability fusion method (Eqn. \ref{eqn:ppf}) leads to a
substantial increase in accuracy: The leave-one-patient-out
cross-validation accuracy increases to 88.3\,\%, at a sensitivity of
86.6\,\% and a specificity of 90.0\,\%. The AUC
is 0.955 (see Fig. \ref{fig:roc}).

\subsubsection*{Transfer Learning on the whole Image}

The transfer learning approach on the maximally sized square image as described in section 
\ref{CNNimageDet} leads to remarkable results, given that a smaller portion of the image is considered for classification. The overall leave-one-patient out cross-validation accuracy is 87.02\,\% at a sensitivity of 90.71\,\% and specificity of 83.80\,\%. As seen in figure \ref{fig:roc}, the area-under-curve is 0.948. 

\vspace{1em}

\vspace{1em}

\begin{figure}[!t]
\centering
\subfloat[ ]{
\includegraphics[height=2.1in]{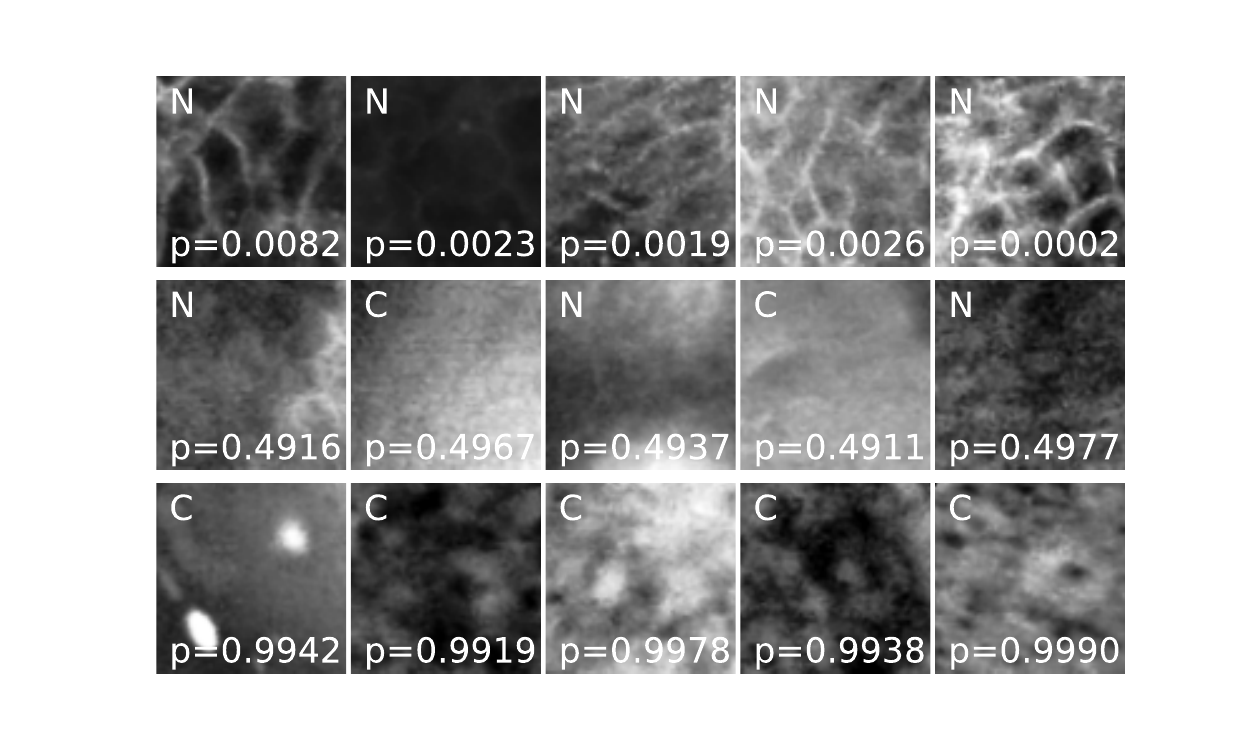}
\label{fig:patchOutcome}
}
\subfloat[ ]{
\includegraphics[height=2.1in]{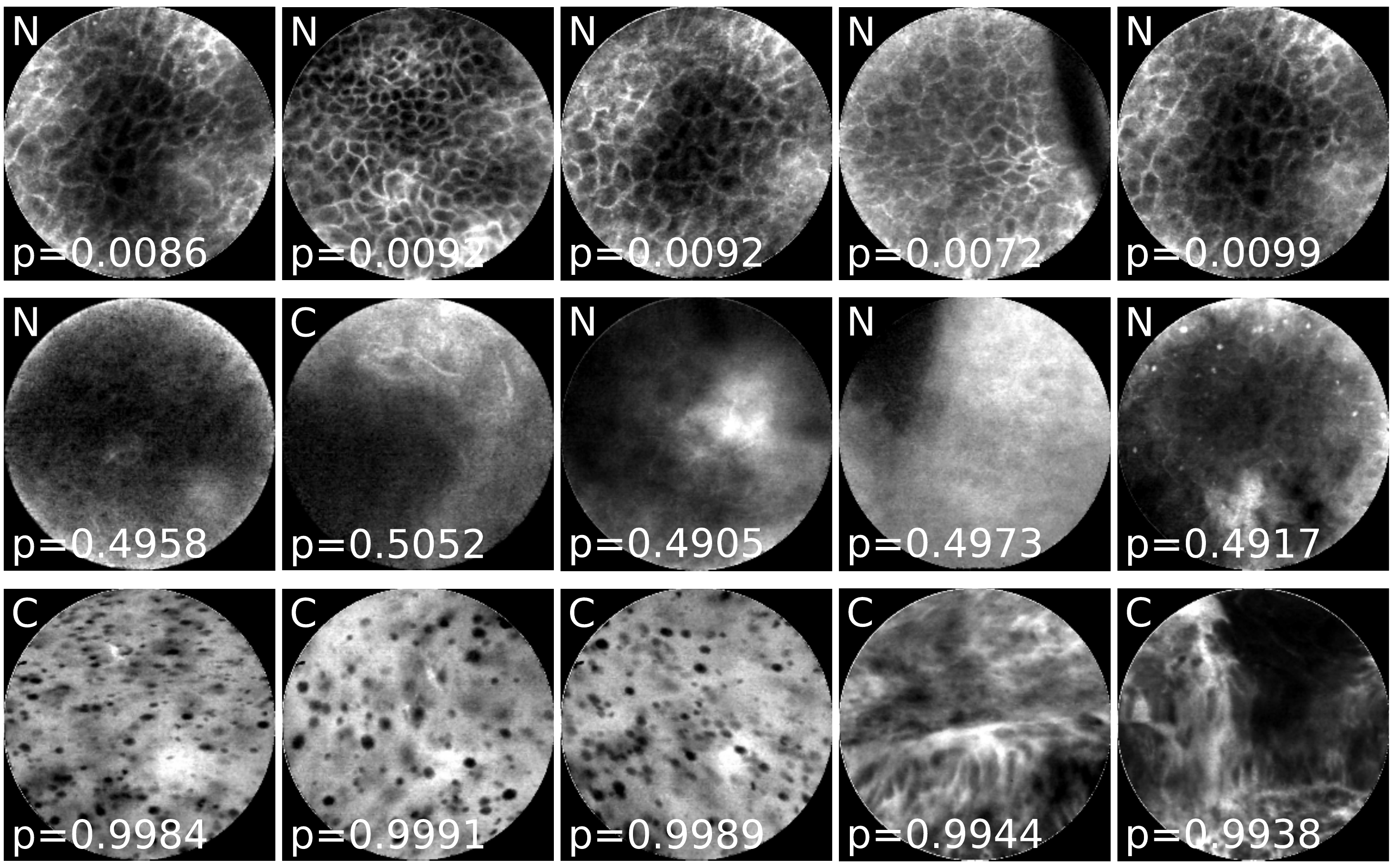}
\label{fig:imageOutcome}
}
\caption{
Randomly selected CNN patch \textbf{(a)} and image \textbf{(b)} classifications with overall high probability for clinically normal and carcinogenic tissue (top and bottom) and
uncertain images (middle). N indicates images representing clinically normal (presumably healthy) tissue, C carcinogenic
  images. The probability given is the a posteriori probability for
  the class carcinogenic.
}

\end{figure}

\section{Discussion}

The textural-feature based methods performed worse as expected on the data set, especially compared to findings of Vo \cite{Vo:2016} and Jaremenko \cite{Jaremenko:2015kh}. This may be explained by the much wider spread image qualities in the data set, which do however represent the clinical use case. The CNN-based method further has a greater inherent structural complexity and may be thus able to cope with different image qualities better.

Over full-image processing, the patch-extraction-based approach however reduces computational complexity significantly, as does the initial rescaling of the image. Reduced complexity will help implementation of a real-time system, and also often contributes to the robustness of a system. 
 
One common criticism about non-feature driven machine learning techniques is that the resulting networks can not be easily interpreted, and it might be unclear what weaknesses and strengths the approach has. Analysis of the patches with high probabilities for one class or the other indicate, that the convolutional neural net primarily looks for cell border structures. To illustrate this, figure \ref{fig:patchOutcome} shows a random pick of highly probable (as of the predicted classification probability) cancerous and clinically normal image patches, as well as images where the classifier was unsure what to choose (probability around chance). The latter would typically have rare occurrences of cell borders or no structure whatsoever. The structures assigned to being cancerous typically show signs of unorganized tissue structure like described by Oetter \textit{et al.}\cite{Oetter:2016cp} or of fluorescein leakage (bright background) or cell clustering (black spots).

Analysis of images detected as being clinically normal with a high probability show typically intact cell border networks (see figure \ref{fig:imageOutcome}, top). The images with an image class probability around chance show, why the image recognition task is sometimes even hard for experts on CLE images (Fig. \ref{fig:imageOutcome}, middle). Although the images shown in the middle row of the figure are all from macroscopical normal epithelium, no clearly organized cell structures can be spotted. For the images where the classifier is sure about being cancerous (Fig. \ref{fig:imageOutcome}, bottom), clear signs of carcinoma can be spotted: Fluorescein leakage as well as unorganized structure are clearly visible.


The principal drawback of rectangular patch extraction of a round
image is, of course, that information at the borders is being
discarded and thus not helping for the classification. We assumed
that all patches of the image are highly representative of the overall
image, and that the border regions can thus be neglected. However, for
images where dysplastic or carcinogenic tissue characteristics are
observable in these parts, inclusion of this data would be helpful.  

Further, besides scaling, no preprocessing was considered at all for the images. Mualla\cite{Mualla:2014ko} and Bier\cite{Bier:2015ju} however showed, that preprocessing can improve detection results in CLE images. 

For a truly automatic assessment of CLE images, it would further be necessary to automatically annotate artifacts. Image quality-based gating should improve overall performance, as already shown for CLE images by Kamen \textit{et al.} \cite{Kamen:2016jw}. Especially for motion and noise artifacts, an accurate grading seems to be possible using textural descriptors. However, for anatomical structures -- typically not relevant for the cancer diagnosis -- this might be much more difficult and might be another interesting task for deep learning.

For the patch probability fusion method, we have chosen a rather simple network topology. However, deep learning does not stop at these basic structures but is aimed at networks with a much higher number of layers. Given enough training material, very deep approaches such as residual neural networks \cite{He:2016ib} could certainly show beneficial behavior on the given field. 

For the transfer learning approach, one main downside is certainly the limitation to a squared image in the middle of the CLE view area, which discards 36\,\% of the available area and thus of the available information. Certainly, an approach tailored towards the round shape of CLE could lead to improvements in detection accuracy here.

One clinically very interesting task is staging of cancerous tissue - from hyperplasia over mild, moderate and severe dysplasia up
to carcinoma in situ\cite{Keith:2013jm}. 
The visual clinical oral examination (COE) can only be seen as a screening method for irregularities and lesions of the oral cavity \cite{Cleveland:2013gc}.
Thus, the gold standard of diagnosis is the histo-pathological diagnosis of the suspicious region. Even though this method allows for a highly accurate identification of malignant oral tissue, a grading of the oral cancer as well as the identification of pre-malignant lesions and cellular dysplasia is still subject to inter-rater as well as intra-rater variabilities. and thus considered as a subjective parameter with rather low reproducibility \cite{Abbey:1995kd}.

Also, from the machine learning point of view this is a challenging task as the occurrence of intermediate stage images is usually rare and the task is much more difficult as differentiation is much harder (even for a human expert viewer). This is also due to the fact that the used CellVizio system has a fixed penetration depth, and thus tissue can only be observed in a defined 2D layer. Yet, carcinogenesis is three-dimensional, and usually originating from deeper layers and thus early stages might not be observable using the imaging system.

One principal remaining question about our data is whether the images marked as clinically normal actually all show physiological tissue.  Since no histopathology was performed to assess the tissue due to ethical restrictions, there might be undiscovered pathologies in the image material. In particular, as oral cancer can be seen as a disease based on the theory of field cancerisation with occurring pre-neoplastic processes all over the oral cavity, a general alteration of the mucosa in this type of patients can not be ruled out. As all patients that were part of the study were diagnosed with HNSCC, it can thus be assumed that the prevalence of physiologically abnormal tissue is increased in this patient group. For a clinically valid procedure, it would be important to include an (age and gender matched) healthy control group, which would yet have to be recorded. Since the intravenous administration of the fluorescent agent comes with a low risk which can not be excluded fully and, above all, taking an invasive biopsy has a remaining risk of complications (such as infection, secondary bleeding or cicatrization), performing this procedure on healthy persons is ethically questionable, and it is unclear to what degree a valid and histo-pathologically correlated acquisition of this data is possible at all.


\section{Summary}
In this work the huge potential of applying deep learning technologies to the field of cancer detection in confocal laser endomicroscopy has been outlined. 

For the first time to our knowledge, image recognition based on Convolutional Neural Networks was successfully applied on CLE images of OSCC. The patch probability fusion method, described in this paper, is shown to significantly outperform the conventional approaches like image texture-based classifiers and even better than transfer learning-based image classification using CNNs.

The automatic identification of cancerous lesions in CLE imaging is a significant step towards a rater-independent, reproducible and real-time diagnostic tool that would surpass the conventional visual and tactile screening for oral cancer.  Moreover, an accurate diagnostic test directly on site would accurately identify and outline high-risk regions that need further investigations by the current gold standard of an invasive biopsy and histopathological assessment.

This study shows a great prospect not only for the CLE imaging of carcinomas in the oral cavity, as squamous cells are omnipresent in the mucosa of the upper aero-digestive and respiratory tract. Further studies have to be conducted to a) to expand the present findings to the more complex task of identification and differentiation of pre-malignant lesions in situ and to b) transfer the findings to further entities of squamous cell carcinomas in the upper aero-digestive tract. Additionally, the task of a real-time identification of OSCC directly on the patient during the screening process will be pursued by our workgroup by optimization of the underlying mathematical algorithms.


\section*{Author contributions and competing financial interests statement}

M.A., C.K and N.Oe wrote the main part of the manuscript. 
The images for this analysis were acquired by N.Oe and C.K.
The toolchain of the new CNN approaches were set up by M.A. 
C.J. provided the toolchain for the textural features-approach. 
M.A., C.J. and N.Oe analyzed the results.
E.R., J.D., F.S. and A.M. provided expertise through intense discussions.
All authors reviewed the manuscript.

The authors declare no competing financial interest.

\end{document}